\documentclass[11pt,a4paper]{article}
\usepackage[hyperref]{acl2020}
\usepackage{times}
\usepackage{latexsym}

\usepackage[utf8]{inputenc}
\usepackage[greek,english]{babel}
\usepackage{comment}
\usepackage{adjustbox}
\usepackage{ragged2e}

\usepackage{microtype}

\aclfinalcopy 

\begin{document}
\selectlanguage{english}

\newcommand{\TODO}[1]{\noindent \textcolor{green}{TODO: #1}}
\newcommand{\khowell}[1]{\noindent \textcolor{blue}{#1}}
\newcommand{\tc}[1]{\noindent \textcolor{cyan}{#1}}
\newcommand{\dom}[1]{\noindent \textcolor{purple}{dom: #1}}

\title{Should Semantic Vector Composition be Explicit? Can it be Linear?\\
\normalsize Semantic Spaces at the Intersection of NLP, Physics, and Cognitive Science (SemSpace2021)
}

\author{Dominic Widdows \\
  LivePerson Inc. \\
  \small{\texttt{dwiddows@liveperson.com}} \\\And
  Kristen Howell \\
  LivePerson Inc. \\
  \small{\texttt{khowell@liveperson.com}} \\\And
  Trevor Cohen \\
  University of Washington \\
  \small{\texttt{cohenta@uw.edu}} \\
  }

\maketitle
\thispagestyle{plain}
\pagestyle{plain}

\vspace{-10cm}

\begin{abstract}
Vector representations have become a central element in semantic language modelling, leading to mathematical overlaps 
with many fields including quantum theory. Compositionality is a core goal for such representations: given representations for
`wet' and `fish', how should the concept `wet fish' be represented?

This position paper surveys this question from two points of view. The first considers the question of whether an 
explicit mathematical representation can be successful using only tools from
within linear algebra, or whether other mathematical tools are needed.
The second considers whether semantic vector composition 
should be explicitly described mathematically, or whether it can be a model-internal side-effect of training a neural network. 

A third and newer question is whether a compositional model can be implemented on a quantum computer. Given the fundamentally 
linear nature of quantum mechanics, we propose that these questions are related, and that this survey may help to highlight
candidate operations for future quantum implementation.

\end{abstract}

\section{Introduction}

Semantic composition has been noted and studied since ancient times, including questions on which parts of language
should be considered atomic, how these are combined to make new meanings, and how explicitly the process of combination can be modelled.\footnote{Early examples are given by Aristotle, such as (\it{De Interpretatione}, Ch IV):

\begin{quote}
    The word `human' has meaning, but does not constitute a proposition, either positive or negative. It is only when other words are added that the whole will form an affirmation or denial.
\end{quote}}

As vectors have come to play a central role in semantic representation, these questions have naturally become asked of semantic vector models.
Early examples include the weighted summation of term vectors into document vectors in information retrieval \cite{salton1975vector} and the
modelling of variable-value bindings using the tensor product\footnote{Familiarity with tensor products is assumed throughout this paper. 
Readers familiar with the linear algebra of vectors but not the multilinear algebra of 
 tensors are encouraged to read the introduction to tensors in \citet[\S 5]{widdows2021quantum}.}
in artificial intelligence \citep{smolensky1990tensor}.
The use of vectors in the context of natural language processing grew from such roots, landmark papers including the introduction
of Latent Semantic Analyis \cite{deerwester-indexing}, where the vectors are created using singular value decomposition, and Word Embeddings \cite{mikolov2013efficient}, where the vectors are created by training a neural net to predict masked tokens.

During the 20th century, logical semantics was also developed, based very much upon the discrete mathematical logic
tradition of \citet{boole-laws} and \citet{frege-arithmetik} rather than the continuous vector spaces that developed
from the works of \citet{hamilton-quaternions} and \citet{grassmann-extension-dup}. Thus, by the beginning of this century,
compositional semantics was developed mathematically, provided frameworks such as Montague semantics for 
connecting the truth value of a sentence with its syntactic form, but provided little insight on how the 
atomic parts themselves should be represented. Good examples in this tradition can be found in \citet{gamut-logic} and \citet{partee-mathematical}. In summary, by the year 2000, there were distributional language 
models with vectors, symbolic models with composition, but little in the way of distributional vector models with composition.

\section{Explicit Composition in Semantic Vector Models}
\label{sec:explicit_composition}

The most standard operations used for composition and comparison in vector model 
information retrieval systems have, for many decades, been the vector sum and cosine similarity 
\citep[see][]{salton1975vector}, and for an introduction, \citet[Ch 5]{widdows-geometry}).
Cosine similarity is normally defined and calculated in terms of the natural scalar product
induced by the coordinate system, i.e.,
\[ \cos(a, b) = \frac{a\cdot b}{\sqrt{(a\cdot a)(b\cdot b)}}. \] 
While the scalar product is a linear operator because $\lambda a \cdot \mu b = \lambda\mu (a\cdot b)$,
cosine similarity is deliberately designed so that $cos(\lambda a, \mu b) = \cos(a, b)$, so that
normalizing or otherwise reweighting vectors does not affect their similarity, which depends only on the angle 
between them.

More sophisticated vector composition in AI was introduced with cognitive models and 
connectionism. The work of \citet{smolensky1990tensor} has already been mentioned, 
and the holographic reduced representations of \citet{plate-hrr} is another widely-cited influence (discussed
again below as part of Vector Symbolic Architectures). While Smolensky's work is often considered
to be AI rather than NLP, the application to language was a key consideration:

\begin{quote}
    Any connectionist model of natural language processing must cope with the
questions of how linguistic structures are represented in connectionist models. \cite[\S 1.2]{smolensky1990tensor}
\end{quote}

The use of more varied mathematical operators to model natural language
operations with vectors accelerated considerably during the first decade of this century. In information retrieval, \citet{rijsbergen-geometry} explored conditional implication and \citet{widdows-negation} 
developed the use of orthogonal projection for negation in vector models. 

Motivated partly by formal similarities with quantum theory, \citet{aerts-semantic} proposed
modelling a sentence $w_1, \ldots, w_n$ with a tensor product $w_1\otimes \ldots \otimes w_n$ in 
the Fock space 
$\bigoplus_{k=1}^\infty V^{\otimes^k}$. The authors noted that comparing the spaces $V^{\otimes^k}$ and $V^{\otimes^j}$ when 
$k \neq j$ 
is a difficulty shared by other frameworks, and of course the prospect of summing to $k=\infty$ is a mathematical
notation that motivates the search for a tractable implementation proposal.

By the middle of the decade,
\citet{clark2007combining} and \citet{widdows-products} proposed and experimented with the use of tensor
products for semantic composition, and the parallelogram rule for relation extraction. 
Such methods were used to obtain strong empirical results
for (intransitive) verb-noun composition by \citet{mitchell2008vector} and for adjective-noun composition by \citet{baroni-nouns}. 
One culmination of this line of research is the survey by \citet{baroni2014frege}, who also
addressed the problem of comparing tensors $V^{\otimes^k}$ and $V^{\otimes^j}$ when 
$k \neq j$. For example, if \citep[as in][]{baroni-nouns}), nouns are represented by vectors and adjectives are
represented by matrices, then the space of matrices is isomorphic to $V\otimes V$ which is not naturally comparable
to $V$, and the authors note \citep[\S3.4]{baroni-nouns}:

\begin{quote}
As a result, Rome and Roman,
Italy and Italian cannot be declared similar, which is counter-intuitive.
Even more counter-intuitively, Roman used as an adjective would not
be comparable to Roman used as a noun.
We think that the best way to solve such apparent paradoxes is
to look, on a case-by-case basis, at the linguistic structures involved,
and to exploit them to develop specific solutions. 
\end{quote}

Another approach is to use a full syntactic parse of a sentence to construct vectors in a sentence space $S$ from
nouns and verbs as constituents in their respective spaces. This features prominently in the model of \citet{coecke-distributional}, 
which has become affectionately known as DisCoCat, 
from `{\it Dis}tributional {\it Co}mpositional {\it Cat}egorical'. The mathematics is at the same time
sophisticated but intuitive: its formal structure relies on pregroup grammars and 
morphisms between compact closed categories, and intuitively, the information from semantic word vectors
flows through a network of tensor products that parallels the syntactic bindings and produces a single
vector in the $S$ space. 

Various papers have demonstrated empirical successes for the DisCoCat and related models. 
\citet{grefenstette2011experimental} were among the first, showing comparable and sometimes improved results 
with those of \citet{mitchell2008vector}. 
By 2014, several tensor composition operations were compared by \citet{milajevs2014evaluating}, and 
\citet{sadrzadeh2018sentence} showed that word, phrase, and sentence entailment can be modelled using vectors
and density matrices. (The use of density matrices to model probability distributions for entailment was pioneered 
partly by \citet[p. 80]{rijsbergen-geometry} and will be discussed further in Section \ref{sec:quantum}.)
Further mathematical tools used in DisCoCat research include {\it copying} a vector $v\in V$ into a tensor in $V\otimes V$
where the coordinates of $v$ become the diagonal entries in the matrix representation of the corresponding tensor, and {\it uncopying}
which takes the diagonal elements of a matrix representing a tensor in $V\otimes V$ and uses these as the coordinates of a vector.
With these additional operators, the tensor algebra becomes more explicitly a {\it Frobenius algebra}.\footnote{Named after Georg Frobenius (1849--1917),
a group theorist who contributed particularly to group representation theory. See \citet{kartsaklis2015compositional} for a thorough presentation.} 
These are used in DisCoCat models  
by \citet{sadrzadeh2014frobenius,sadrzadeh2014frobenius2} to represent relative and then possessive pronoun attachments (for example, representing the
affect of the phrase ``that chased the mouse'' as part of the phrase ``The cat that chased the mouse''). The method involves
detailed tracking of syntactic types and their bindings, and certainly follows the suggestion from \citet{baroni-nouns} to look
at linguistic structures on a case-by-case basis.

There are practical concerns with tensor product formalisms. The lack of any
isomorphism between $V^{\otimes^k}$ and $V^{\otimes^j}$ when $k \neq j$ and $\dim V > 1$ has already been noted, along with the 
difficulty this poses for comparing elements of each for similarity. Also, there
is an obvious computational scaling problem: if $V$ has dimension $n$, then $V^{\otimes^k}$ has dimension $n^k$,
which leads to exponential memory consumption with classical memory registers. Taking the example of relative pronouns in 
the DisCoCat models of \citet{sadrzadeh2014frobenius} --- these are represented as rank-4 tensors in spaces such as $N\otimes S\otimes N\otimes N$
and variants thereof, and if the basic noun space $N$ and sentence space $S$ have dimension 300 (a relatively standard number used e.g., by FastText vectors)
then the relative pronouns would have dimension 8.1 billion. If this was represented densely and the coordinates are 4-byte floating point numbers, then 
just representing one pronoun would require over 30GB of memory, which is intractable even by today's cloud computing standards.

The development of Vector Symbolic Architectures (VSAs) \cite{gayler_vector_2004_dupe} was partly motivated by these concerns.
VSAs grew from the holographic reduced representations of \citet{plate-hrr}: notable works in this intersection of cognitive science and artificial intelligence include those of \citet{eliasmith-howto}
and \citet{kanerva_hyperdimensional_2009-dup}. At its core, a VSA is a vector space with an addition operator and
a scalar product for computing similarity, along with a multiplication or {\it binding} operator (sometimes written as $\ast$, or $\otimes$ like the tensor product) 
which  takes a product of two vectors and returns a new vector that it typically {\it not} similar to either of its 
inputs, so that $(a\ast b) \cdot a$ is small, but which is `approximately reversible' --- so there is an approximate inverse operator $\oslash$ where $(a\ast b) \oslash b$ is close to $a$.\footnote{This also explains why
it is tempting to reuse or abuse the tensor product notation and use the symbol $\otimes$ for binding and $\oslash$
for the inverse or release operator, as in \citet{widdows2014reasoning}.} The term `binding' was used partly for
continuity with the role-filler binding of \citet{smolensky1990tensor}.\footnote{The requirement that the binding $a\ast b$
be dissimilar to both $a$ and $b$ makes the Frobenius uncopying of \citet{kartsaklis2015compositional} operator unsuitable for a VSA, 
because the coordinates of $v\ast w$ are the products of the 
corresponding coordinates in $v$ and $w$, which typically makes the scalar product with either factor quite large. This is however
a rather shallow observation, and the relationship between VSAs and Frobenius algebras may be a fruitful topic to investigate more thoroughly.}

The VSA community has tended to avoid the full tensor product, for the reasons given above. In order to be directly
comparable, it is desirable that $a\ast b$ should be a vector in the space $V$. \citet{plate-hrr} thoroughly 
explored the use of {\it circular correlation} and {\it circular convolution} for these operations, which involves
summing the elements of the outer product matrix along diagonals. This works as a method to map $V\otimes V$ 
back to $V$, though the mapping is of course basis dependent. Partly to optimize the binding operation to $O(n)$ time, \citet[Ch 5]{plate-hrr} introduces {\it circular vectors}, whose coordinates are unit complex numbers. 
There is some stretching of terminology here, because the circle group $U(1)$ of unit complex numbers is not, 
strictly speaking, a vector space. Circular vectors are added by adding their rectangular coordinates in a linear fashion, and then normalizing back to the unit circle by discarding the magnitude, which Plate notes is 
not an associative operation. \citet{kanerva_hyperdimensional_2009-dup} departs perhaps
even further from the vector space mainstream, using binary-valued vectors throughout, with binding implemented as pairwise exclusive OR (XOR).

VSA binding operations have been used for composition of semantic vector representations, both during the training process to generate composite vector representations of term or character n-grams, or semantic units such as predicate-argument pairs or syntactic dependencies, that are then further assembled to construct representations of larger units \cite{jones-holographic, kachergis2011orbeagle,cohen2017embedding, paullada2020improving}; and to compose larger units from pretrained semantic vectors for downstream machine learning tasks \cite{fishbein2008integrating,mower2016classification}. However, a concern with several of the 
standard VSA binding operators for the representation of sequences in particular is that they are commutative in nature: $x \ast y = y \ast x$. 
To address this concern, permutations of vector coordinates have been applied across a range of VSA implementations to break the commutative property of the binding operator, for example by permuting the second vector in sequence such that $\overrightarrow{wet} \ast \prod (\overrightarrow{fish})$ and $\overrightarrow{fish} \ast \prod (\overrightarrow{wet})$ result in different vectors \cite[p. 121]{kanerva_hyperdimensional_2009-dup,plate-hrr}.

Thanks to their general nature and computational simplicity, 
permutations have been used for several other encoding and composition experiments.
The use of permutations to encode positional information into word vector representations was introduced by \citet{sahlgren_permutations_2008_dupe}. In this work a permutation (coordinate shuffling) operator was used to rearrange vector components during the course of training, with a different random permutation assigned to each sliding window position such that a context vector would be encoded differently depending upon its position relative to a focus term of interest. A subsequent evaluation of this method showed advantages in performance over the BEAGLE model \cite{jones-holographic}, which uses circular convolutions to compose representations of word n-grams, on a range of intrinsic evaluation tasks --- however these advantages were primarily attributable to the permutation-based approach's ability to scale to a larger training corpus \cite{recchia2015encoding}. Random permutations have also been used to encode semantic relations \cite{cohen2009predication} and syntactic dependencies \cite{basile2011encoding} into distributional models.  

In high-dimensional space, the application of two different random permutations to the same vector has a high probability of producing vectors that are close-to-orthogonal to one another \cite{sahlgren_permutations_2008_dupe}. A more recent development has involved deliberately constructing `graded' permutations by randomly permuting part of a parent permutation \cite{cohen2018bringing}. When this process is repeated iteratively, it results in a set of permutations that when applied to the same vector will produce a result with similarity to the parent vector that decreases in an ordinal fashion. This permits the encoding of proximity rather than position, in such a way that words in proximal positions within a sliding window will be similarly but not identically encoded. The resulting proximity-based encodings have shown advantages over comparable encodings that are based on absolute position (at word and sentence level) or are position-agnostic (at word and character level) across a range of evaluations \cite{cohen2018bringing,schubert2020reading,kelly2020sentence}. 


Note that coordinate permutations are all Euclidean transformations: odd permutations are reflections, and even permutations are rotations. Thus all permutation
operations are also linear.


This survey of explicit composition in semantic vector models is not exhaustive, but gives some idea of the range of linear
and nonlinear operations.

\section{Compositional Semantics in Neural Networks}

During the past decade, many of the most successful and well-known advances in semantic vector
representations have been developed using neural networks.\footnote{An introduction to this huge topic
is beyond the scope of this paper. Unfamiliar readers are encouraged to start with a general
survey such as that of \citet{geron2019hands}, Chapter 16 being particularly relevant to the discussion here.}
In general, such networks are trained with some objective function designed to maximize the probability
of predicting a given word, sentence, or group of characters in a given context.
Various related results such as those of \citet{scarselli1998universal} are known to demonstrate that,
given enough computational resources and training data, neural networks can be used to approximate 
any example from large classes of functions. If these target functions are nonlinear, this cannot be done
with a network of entirely linear operations, because the composition of two linear maps is another
linear map --- ``The hidden units should be nonlinear because multiple layers of linear units can only produce linear functions.'' \cite[\S 13.5]{wichert:principles}. Thus, part of the appeal of neural networks is that they are {\it not} bound by linearity:
though often at considerable computational cost.


The skip gram with negative sampling method was introduced by \citet{mikolov2013efficient}, implementations
including the word2vec\footnote{\url{https://pypi.org/project/word2vec/}}
package from Google and the FastText package from Facebook.\footnote{\url{https://fasttext.cc/}}
The objective function is analyzed more thoroughly by \citet{goldberg2014word2vec}, and takes the form:

\begin{eqnarray}
\label{sgns_optimization_objective}
\sum_{(w,c) \in D}\log{\sigma} (\overrightarrow{w} \cdot \overrightarrow{c}) + \sum_{(w,\neg c) \in D'} {\log{\sigma}} ( - \overrightarrow{w} \cdot \overrightarrow{\neg c})  \nonumber
\end{eqnarray}

Here $w$ is a word, $c$ is a context feature (such as a nearby word), $D$ represents observed term/context pairs in the document collection, $D'$ represents randomly drawn counterexamples, and $\overrightarrow{w}$ and $\overrightarrow{c}$ are word and context vectors (input and output weights of the network, respectively). $\sigma$ is the sigmoid function,  $\frac{1}{1 + e^{-x}}$.
The mathematical structure here is in the family of logistic and softmax functions --- the interaction between
the word and context vectors involves exponential / logarithmic concepts, not just linear operations.

There have been efforts to incorporate syntactic information explicitly in the training process of neural network models. 
In the specific case of adjectives, \citet{maillard2015learning} use the skip gram technique to create matrices for
adjectives following the pattern of \citet{baroni-nouns} discussed in Section \ref{sec:explicit_composition}.
The most recent culmination of this work is its adaptation to cover a much more comprehensive collection of categorial types by
\citet{wijnholds2020representation}.
Another early example comes from \citealt{socher2012matrix}, who train a Recursive Neural Network where each node in a syntactic parse tree becomes represented
by a matrix that operates on a pair of inputs. Research on tree-structured LSTMs \citep[see inter alia][]{tai2015improved, maillard2019jointly} leverages syntactic parse trees in the input and composes its hidden state using an arbitrary number of child nodes, as represented in the syntax tree.
Syntax-BERT \citep{bai2021syntax} uses syntactic parses to generate masks that reflect different aspects of tree structure (parent, child, sibling).
KERMIT \citep{zanzotto2020kermit} uses compositional structure explicitly by embedding syntactic subtrees in the representation space.
In both cases, the use of explicit compositional syntactic structure leads to a boost in performance on various semantic tasks.

In KERMIT, the embedding of trees and (recursively) their subtrees follows a well-developed line of research on representing discrete structures 
as vectors, in particular combining circular convolution and permutations to introduce {\it shuffled circular convolution} \cite{ferrone2014towards}.
Even when combined in a recursive sum over constituents called a Distributed Tree Kernel operation, this is still a sum of linear inputs, so this form of 
composition is still linear throughout. In such methods, the result may be a collection of related
linear operators representing explicit syntactic bindings, but the training method is typically not linear due to the activation functions.

What these neural net methods and the models described in the previous section have in common is that
they encode some {\it explicit} compositional structure: a weighted sum of word or character n-grams, a
role / value binding, or a relationship in a grammatical parse tree. This raises the question: can neural 
language models go beyond the bag-of-words drawbacks and encode more order-dependent
language structures without using traditional logical compositional machinery? 

A recent and comprehensive survey of this topic is provided by \citet{hupkes2020compositionality}.
This work provides a valuable survey of the field to date, and conducts experiments
with compositional behavior on artificial datasets designed to demonstrate various aspects of
compositionality, such as productivity (can larger unseen sequences be produced?) and substitutivity (are outputs the same when synonymous tokens are switched?). This systematic approach to breaking compositionality into many tasks is a useful guide in itself.

Since then, attention-based networks were developed and have come to the forefront of the field \cite{vaswani2017attention}. 
The attention mechanism is designed to learn when pairs of inputs depend crucially on one another, a capability that has demonstrably improved machine translation
by making sure that the translated output represents all of the given input 
even when their word-orders do not correspond exactly. 
The `scaled dot-product attention' used by \citet{vaswani2017attention} 
for computing the attention between a pair of constituents uses softmax normalization, another nonlinear
operation.

The use of attention mechanisms has led to rapid advances in the field,
including the contextualized BERT \cite{devlin2018bert} and ELMo \cite{peters2018deep} models. For example, the ELMo model reports
good results on traditional NLP tasks including question answering, 
coreference resolution, semantic role labelling, and part-of-speech tagging, 
and the authors speculate that this success is due to the model's different neural-network
layers implicitly representing several different kinds of linguistic structure. This idea is further investigated by \citet{hewitt2019structural} and \citet{jawahar2019does}, who probe BERT and ELMo models to find evidence that syntactic structure is implicitly encoded in their vector representations.
The survey and experiments of \citet{hupkes2020compositionality} evaluate three such neural networks on 
a range of tasks related to composition, 
concluding that each network has strengths and weaknesses, that the results are a stepping stone rather than an endpoint, 
and that developing consensus around how such tasks should be designed, tested and shared is a crucial task in itself. 

At the time of writing, such systems are contributors to answering a very open research question: do neural networks need 
extra linguistic information in their input to properly work with language, or
can they actually {\it recover} such information as a byproduct of 
training on raw text input? For example, a DisCoCat model requires parsed sentences
as input --- so if another system performed as well without requiring grammatical sentences as input 
and the support of a distinct parsing component in the implementation pipeline, 
that would be preferable in most production applications. 
(Running a parser is a requirement than today can often be satisfied, albeit with an implementational and computational cost. 
Requiring users to type only grammatical input is a requirement that cannot typically be met at all.)
At the same time, does performance on the current NLP tasks used for evaluation directly indicate semantic composition at play?
If the performance of a model without linguistic information in the input is up to par, would the internal operations of such an implicit model be 
largely inscrutable, or 
can we describe the way meaningful units are composed into larger meaningful structures explicitly?

Tensor networks are one of the possible mathematical answers to this question, and continue
to build upon Smolensky's introduction
of tensors to AI. For example \citet{mccoy2020tensor} present evidence that the sequence-composition
effects of Recurrent Neural Networks (RNNs) can be approximated by Tensor Product Decomposition Networks, 
at least in cases where using this structure provides measurable benefits over bag-of-words models. 
It has also been shown that Tensor Product Networks
can be used to construct an attention mechanism from which grammatical structure can be recovered
by unbinding role-filler tensor compositions \cite{huang2019attentive}. 

While there are many more networks that could be examined in a survey like this, those described in this section illustrate that neural networks have been used to improve results with many NLP tasks, and the training
of such networks often crucially depends on nonlinear operations on vectors. Furthermore, while tensor networks have been developed as a proposed family of techniques for understanding and
exploiting compositional structures more explicitly, in some of the most 
state-of-the-art models, relationships between such operations to more traditional 
semantic composition are often absent or at least not well-understood.

\section{Operators from Quantum Models}
\label{sec:quantum}

Mathematical correspondences between vector models for semantics and quantum theory have been recognized
for some years \cite{rijsbergen-geometry}, and are surveyed by \citet{widdows2021quantum}. 
The advent of practical quantum computing makes these correspondences especially interesting, and 
constructs from quantum theory have been used increasingly deliberately in NLP.
In quantum computing, tensor products no longer incur quadratic costs: instead, 
the tensor product $A\otimes B$ is the natural mathematical representation of the physical state 
that arises when systems in states $A$ and $B$ are allowed to interact. Heightened interest in 
quantum computing and quantum structures in general has led to specific semantic contributions already.

Mathematically, there is a historic relationship between linearity and quantum mechanics: 
the principle of superposition guarantees that for any state $A$, the vector $c A$ corresponds to the same
physical state for any complex number $c$ \cite[\S 5]{dirac-quantum-dup}.\footnote{This itself could
lead to a mathematical discussion --- the magnitude of a state vector in quantum mechanics is ignored,
just like cosine similarity ignores the scale factors of the scalar product, and its resilience to
scale factors makes the cosine similarity technically {\it not} a linear operator.} Hence the question of
whether a compositional operator is linear or not is particularly relevant when we consider the practicality of
implementation on a quantum computer.\footnote{The dependence of quantum computing on linearity should not go unquestioned ---
for example, the use of quantum harmonic oscillators rather than qubits 
has been proposed as a way to incorporate nonlinearity into quantum hardware by \citet{goto2016bifurcation}.}

Many developments have followed from the DisCoCat framework, whose mathematical structure is closely related
to quantum mechanics through category theory \cite{coecke2018uniqueness}. As of 2021, the tensor network components of 
two DisCoCat models have even been implemented successfully on a quantum computer \cite{lorenz2021qnlp},
and there are proposals for how to implement the syntactic parse on quantum hardware as well \cite{wiebe2019quantum,bausch2021quantum}.
Of particular semantic and mathematical interest, 
topics such as hyponymy \cite{bankova2019graded} and negation \cite{lewis2020towards} have been investigated, 
using density matrices and positive operator-valued measures, which are mathematical generalizations of 
state vectors and projection operators that enable the theory to describe systems that are not in `pure' states.
Density matrices have also been used to model sentence entailment \cite{sadrzadeh2018sentence} and recently lexical ambiguity \cite{meyer2020modelling}.

A comprehensive presentation of the use of density matrices to model joint probability distributions is given by \citet{bradley2020interface}. This work deliberately takes a quantum probability framework and applies it to
language modelling, by way of the lattice structures of Formal Concept Analysis \cite{ganter-formal}. 
This work uses the {\it partial trace} of density operators (which are tensors in $V\otimes V$) to project
tensors in $V\otimes V$ to vectors in $V$. This is analogous to summing the rows or columns of a two-variable 
joint distribution to get a single-variable marginal distribution. This captures interference and overlap between
the initial concepts, and in a framework such as DisCoCat, this might be used to model transitive verb-noun composition
\citep[as in][and others]{grefenstette2011experimental,sadrzadeh2018sentence}.

Another mathematical development is the quantum Procrustes alignment method of \citet{lloyd2020quantum}, where 
Procrustes alignment refers to the challenge of mapping one vector space into another preserving relationships
as closely as possible. Procrustes techniques have been used to align multilingual FastText word vectors \cite{joulin2018loss}, 
and it is possible that one day these methods may be combined to give faster and more noise-tolerant multilingual concept
alignment.

This again is not a complete survey, but we hope it demonstrates that the interplay between quantum theory, semantic vector
composition, and practical implementation has much to offer, and that work in this area is accelerating.


\begin{table*}
\centering
\caption{Survey Summary of Mathematical Methods Used for Semantic Vector Composition}
\label{operator-table}
\RaggedRight
\renewcommand{\arraystretch}{1.4}

\begin{adjustbox}{width=\textwidth}
\begin{small}
\begin{tabular}{|p{2.7cm}|p{3cm}|p{2.5cm}|p{2.5cm}|p{2cm}|p{2.2cm}|p{2.5cm}|}
\hline
\bf{Mathematical Method} & \bf{Use Case} & \bf{Inputs} & \bf{Outputs} & {\bf Explicitness} & \bf{Linearity} & \bf{Described By} \\
\hline
Vector sum & Document retrieval and many others & Word vectors & Document vectors & Explicit & Linear & Widespread including \citet{salton1975vector} \\
Scalar product & Similarity scoring & Any vector & Any vector & Explicit & Linear & Various \\
Cosine similarity & Similarity scoring & Any vector & Any vector & Explicit & Nonlinear --- deliberately ignores magnitude & Various \\
Tensor product & Role-filler binding & Variable name & Variable value & Explicit & Linear & \citet{smolensky1990tensor} \\
Circular vector sum & Holographic reduced representations & Circular vectors & Circular vectors & Explicit & Nonlinear, and non-associative & \citet{plate-hrr} \\
(Stalnaker) Conditional & Implication & Vector / subspace representing propositions & Subspace representing truth conditions & Explicit & Linear (though for subspaces it doesn't matter) & \citet[Ch 5]{rijsbergen-geometry} \\
Orthogonal projection & Negation & Vector or subspace & Vector & Explicit & Linear & \citet{widdows-negation} \\
Sum of subspaces & Disjunction & Vectors or subspaces & Subspace & Explicit & Linear & \citet{widdows-negation} \\
Parallelogram rule & Proportional analogy & Three vectors & Fourth vector & Explicit & Linear & Various incl. \citet{widdows-products}, \citet{mikolov2013efficient} \\
Tensor product & Word vector composition & Word vector & Sentence-fragment tensor & Explicit & Linear & Various since \citet{aerts-semantic,clark2007combining} \\
Tensor and monoidal product & Parallel semantic, syntactic composition & (vector, syntactic type) pairs & Sentence vectors & Explicit & Linear & Various since \citet{coecke-distributional} \\
Matrix multiplication & Adjective / noun composition & Matrix and vector & Vector & Explicit & Linear & \citet{baroni-nouns} \\
Circular convolution & Vector binding & VSA vector & VSA vector & Explicit & Sometimes & \citet{plate-hrr}, options in \citet{widdows2014reasoning} \\
Binary XOR & Binary vector binding & VSA vector & VSA vector & Explicit & Binary vectors warrant more discussion! & \citet{kanerva_hyperdimensional_2009-dup} \\
Permutation of coordinates & Non-additive composition & Vector & Vector & Explicit, though often random & Linear (because rotation or reflection) & \citet{sahlgren_permutations_2008_dupe} and various \\
Skipgram objective & Vector interation in training & Word and context vector & Update to both & Explicit though internal & Nonlinear & \citet{mikolov2013efficient} \\
$\tanh$, Sigmoid, ReLU, Softmax, etc. & Activation functions in neural networks & Input weights & Output weights & Typically implicit & 
Nonlinear & Many including \citet[Ch 10]{geron2019hands} \\ 
Scaled dot-product attention & Learning pairwise dependence & Vectors & Updated vectors & Typically internal & Nonlinear & \citet{vaswani2017attention} \\
Distributed tree kernel / shuffled circular convolution & Embedding syntactic tree in vector space & Parse tree & Sentence vector & Explicit & Linear & \cite{ferrone2014towards} \\
Density matrices, POVMs & More general distributions over vector spaces, e.g., representing categories, implication & Several, e.g., superpositions of pairs of vectors & Projected vectors and / or probabilities & Often explicit & Linear & \citet{rijsbergen-geometry,sadrzadeh2018sentence,lewis2020towards,bradley2020interface} \\
Procrustes alignment & Aligning vector models $U$ and $V$ & Pairs of source, target vectors & Linear mapping from $U$ to $V$ & Explicit &
Linear & \citet{bojanowski2017enriching,lloyd2020quantum} \\

\hline
\end{tabular}
\end{small}
\end{adjustbox}
\end{table*}

\section{Summary, Conclusion, and Future Work}

This paper has surveyed vector composition techniques used for aspects of semantic composition
in explicit linguistic models, neural networks, and quantum models, while acknowledging that
these areas overlap. The operations considered are gathered and summarized in 
Table \ref{operator-table}.

Some of the most successful neural network models to date have used operations that are nonlinear and implicit.
Though models such as BERT and ELMo have performed exceptionally well on several benchmark tasks, they 
are famously difficult to explain and computationally expensive. Therefore, scientific researchers and commercial user-facing enterprises have good reason to be impressed, 
but still to seek alternatives that are clearer and cheaper. 
At the same time, progress in quantum computing raises the possibility that the practical cost
of different mathematical operations may be considerably revised over the coming year.
For example, if the expense of tensor products becomes linear rather than quadratic,
tensor networks may find a position at the forefront of `neural quantum computing'.

In addition, there is emerging evidence that such models can be augmented by approaches that draw on structured semantic knowledge  \cite{michalopoulos2020umlsbert,colon2021combining}, suggesting the combination of implicit and explicit approaches to semantic composition as a fruitful area for future methodological research. We hope that this approach of surveying and comparing the semantic, mathematical and computational elements of various vector operations will serve as a guide to territory yet to be explored at the intersection of compositional operators and vector representations of language. 

\section{Acknowledgements}

The authors would like to thank the anonymous reviewers for helping to improve the coverage of this survey,
and the conference organizers for allowing extra pages to accommodate these additions.

\bibliography{anthology,acl2020,semanticvectors}
\bibliographystyle{agsm}

\end{document}